\documentclass{article}
\usepackage{titlesec}
\titleformat{\section}[block]{\normalfont\Large\bfseries}{\thesection}{1em}{}  

\makeatletter
\def\@afterheading{\par\@afterindenttrue}
\makeatother
\usepackage{ijcai25}

\usepackage{times}
\usepackage{soul}
\usepackage{url}
\usepackage[hidelinks]{hyperref}
\usepackage[utf8]{inputenc}
\usepackage[small]{caption}
\usepackage{graphicx}
\usepackage{amsmath}
\usepackage{amsthm}
\usepackage{booktabs}
\usepackage{algorithm2e}
\usepackage[switch]{lineno}
\usepackage{etoolbox}  
\usepackage{color}
\usepackage{graphicx} 
\usepackage{amssymb}
\usepackage{subfig}
\usepackage{verbatim}

\title{HySurvPred: Multimodal Hyperbolic Embedding with Angle-Aware Hierarchical Contrastive Learning and Uncertainty Constraints for Survival Prediction}
\author{
    Jiaqi Yang$^{1,2,5}$, 
    Wenting Chen$^{3}$, 
    Xiaohan Xing$^{4}$, 
    Sean He$^{2}$, 
    Xiaoling Luo$^{1}$, 
    Xinheng Lyu$^{1,2}$, 
    Linlin Shen$^{1}$, 
    Guoping Qiu$^{2,5}$
    \\
    $^{1}$ Shenzhen University, 
    $^{2}$ University of Nottingham, Ningbo,
    $^{3}$ City University of Hong Kong,
    $^{4}$ Stanford University,
    $^{5}$ University of Nottingham, UK 
}

\begin{document}

\maketitle

\begin{abstract}
Multimodal learning that integrates histopathology images and genomic data holds great promise for cancer survival prediction. However, existing methods face key limitations: 1) They rely on multimodal mapping and metrics in Euclidean space,  which cannot fully capture the hierarchical structures in histopathology (among patches from different resolutions) and genomics data (from genes to pathways). 2) They discretize survival time into independent risk intervals, which ignores its continuous and ordinal nature, and fails to achieve effective optimization. 3) They treat censorship as a binary indicator, excluding censored samples from model optimization and not making full use of them. 
To address these challenges, we propose HySurvPred, a novel framework for survival prediction that integrates three key modules: \textbf{Multimodal Hyperbolic Mapping} (MHM), \textbf{Angle-aware Ranking-based Contrastive Loss} (ARCL) and \textbf{Censor-Conditioned Uncertainty Constraint (CUC)}. 
Instead of relying on Euclidean space, we design the MHM module to explore the inherent hierarchical structures within each modality in hyperbolic space. 
To better integrate multimodal features in hyperbolic space, we introduce the ARCL module, which uses ranking-based contrastive learning to preserve the ordinal nature of survival time, along with the CUC module to fully explore the censored data.
Extensive experiments demonstrate that our method outperforms state-of-the-art methods on five benchmark datasets. 
Source code is to be released.
\end{abstract}
\begin{figure}[!ht]
\centering
\includegraphics[width=1.0\linewidth]{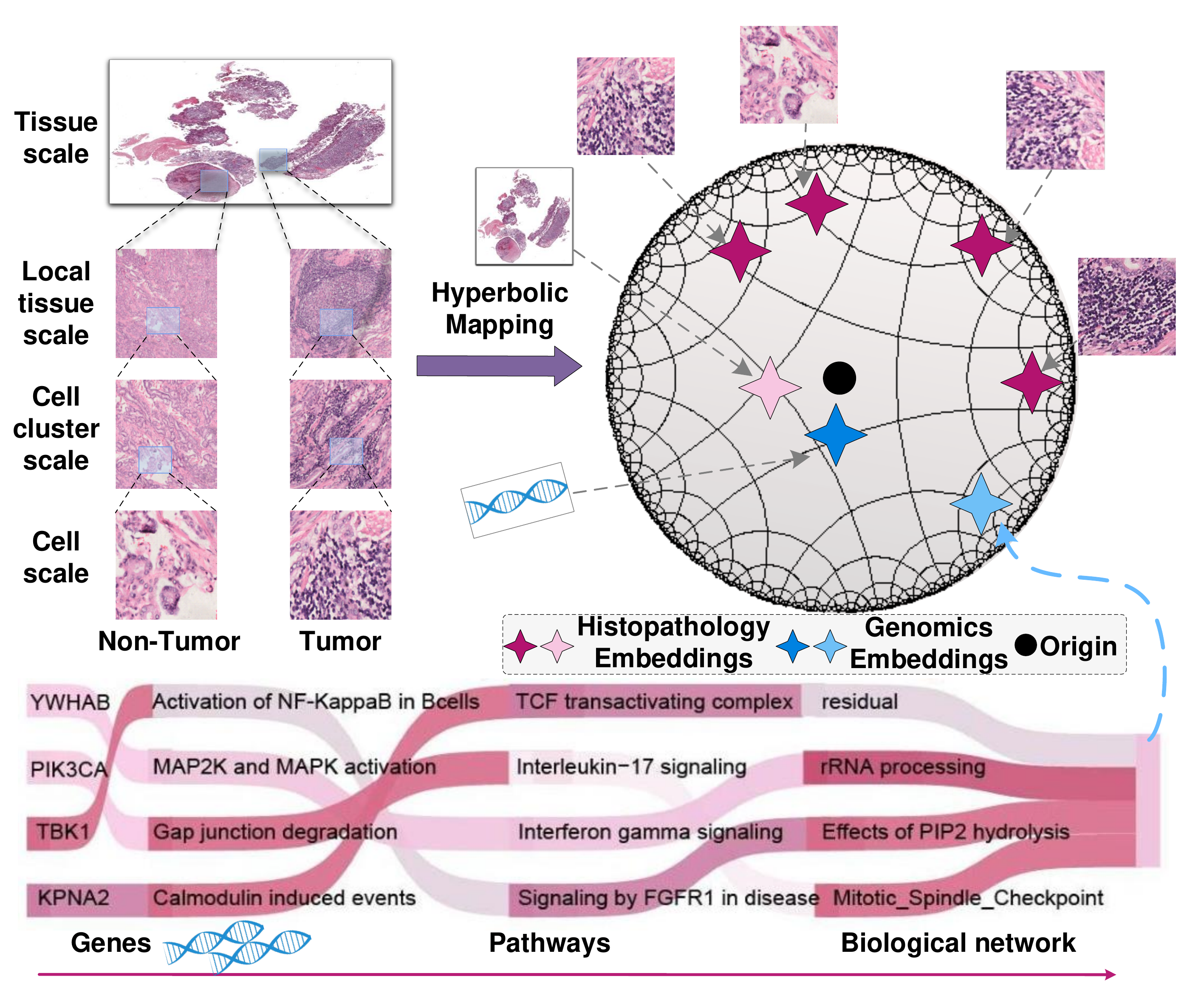}
\caption{Our HySurvPred extracts original features from two highly structural modalities (i.e., histopathology slides and genomic data), and maps them into hyperbolic space for feature fusion.}
\label{Figure1}
\end{figure}

\section{Introduction}
\noindent
Survival prediction in cancer prognosis focuses on estimating death risk and identifying biomarkers indicative of disease progression \cite{song2024morphological,song2024analysis,chen2022pan}. To accurately predict survival outcomes, it is essential to incorporate information from two key modalities, i.e. histopathology and genomics. Histopathology, represented as gigapixel Whole-Slide Images (WSIs), provides detailed spatial depictions of tissue morphology, including tumor microenvironment (TME) features like fibroblasts and immune cells, which are strongly associated with survival outcomes \cite{zhu2017wsisa,oya2020tumor,kuroda2021tumor}. On the other hand, transcriptomics data from bulk RNA sequencing offer insights into gene expression profiles, which are governed by gene networks, regulatory pathways, and interactions between various molecular components \cite{wang2021gpdbn}. The information from these two modalities is complementary, and their fusion is beneficial for revealing TME features and predicting survival outcomes. To achieve this, multimodal approaches are increasingly being adopted, combining qualitative morphological data from histology with quantitative molecular profiles from genomics \cite{zuo2022identify}. Multimodal approaches, integrating histology and transcriptomics, have shown promise due to their complementary strengths \cite{xu2023multimodal,chen2021multimodal}.

Despite the gains made by multimodal methods over unimodal approaches, current methods still fail to fully exploit the valuable information from both modalities. There are three limitations: 1) Current multimodal methods struggle to fully explore the hierarchical information in both data modalities. As shown in Fig. \ref{Figure1}, WSIs exhibit a hierarchy of \textit{tissue} $\rightarrow$ \textit{cell clusters} $\rightarrow$ \textit{cells}, and genomics data forms a hierarchy of \textit{biological network} $\rightarrow$ \textit{pathways} $\rightarrow$ \textit{genes} \cite{qiu2023deep}. These hierarchical structures can enhance feature representation and survival prediction \cite{zhang2020evaluation,zhou2021hyperbolic}.
However, most existing methods rely on Euclidean space \cite{xu2023multimodal} for representation learning, which assumes linearity and uniformity, making it unsuitable for capturing these hierarchical and non-linear relationships. 
2) Current survival prediction methods typically treat survival time as a classification task by dividing it into discrete risk intervals. This discretization approach fails to capture the true nature of survival data, which is inherently continuous and ordinal. While the time intervals naturally progress from low to high risk, traditional classification methods treat these categories as independent labels, losing both the continuous temporal information and the ordinal relationships between risk levels. Although reformulating survival prediction as a regression task could potentially address these issues by directly modeling the continuous survival time, such models often face convergence difficulties during optimization. 3) Existing methods primarily treat censorship as a binary indicator (1 or 0) of uncertainty in survival risk prediction, using it to re-weight the negative log-likelihood loss \cite{zadeh2020bias}. When a data sample's censorship indicator is 0, its corresponding loss becomes zero, effectively excluding it from model optimization. This is particularly concerning since censored samples constitute approximately 45\% of the dataset TCGA-LUAD \cite{9710773}. Despite being censored, these samples could still provide valuable information for learning survival predictions. Thus, it is necessary to explore the hierarchical information in the multimodal data, consider ordinal relationships among risk levels, and exploit the valuable censored data.

\begin{figure*}[t!]
\centering
\includegraphics[width=1.0\linewidth]{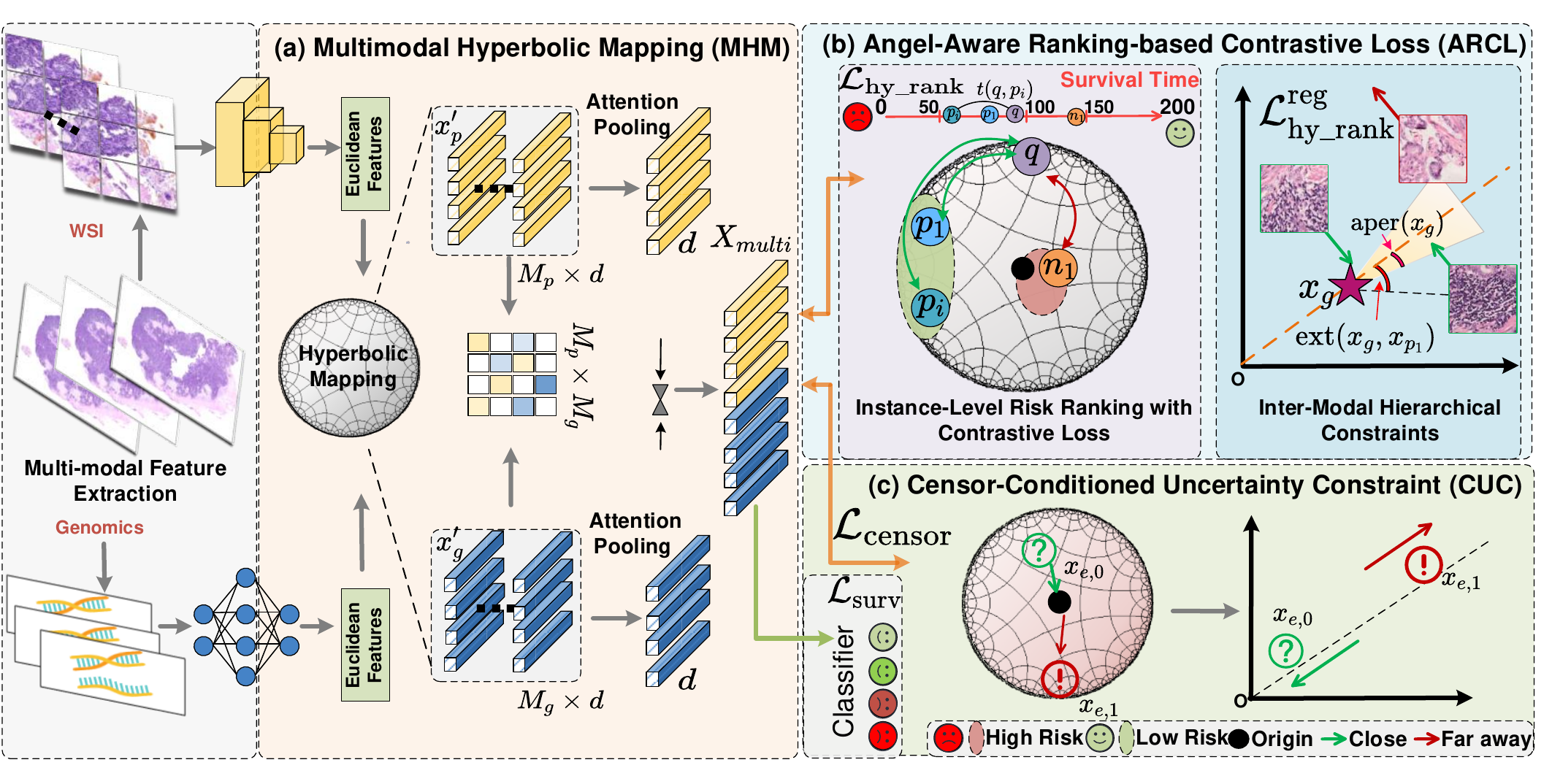}
\caption{Overview of our HySurvPred, a novel framework for survival prediction, with a Multimodal Hyperbolic Mapping (MHM) to explore the inherent hierarchical structures within modality in hyperbolic space, an Angle-aware Ranking-based Contrastive Loss (ARCL) to preserve the ordinal nature of survival time and a Censor-Conditioned Uncertainty Constraint (CUC) to fully explore the censored data for optimization. 
}
\label{fig:framework}
\end{figure*}

To address these issues, we propose a HySurvPred model that explores hierarchical intra-modal structures and inter-modal relationships in histopathology and genomics data.
Different from the commonly used Euclidean space, the hyperbolic space naturally represents hierarchical relationships through exponential growth, making it well-suited to capture the inherent hierarchies in complex biological data \cite{zhang2020evaluation,zhou2021hyperbolic}. Thus, we propose a \textit{Multimodal Hyperbolic Mapping} (MHM) module to effectively capture both local and global relationships in histopathology slides and genomics data, preserving hierarchies within and across modalities. 
Next, to enhance the alignment of multimodal features in hyperbolic space and preserve hierarchical information, while optimizing the relative ordering of samples through ranking-based contrastive learning to account for the continuity of survival time, we introduce an \textit{Angle-aware Ranking-based Contrastive Loss} (ARCL), as shown in Fig. \ref{fig:framework}.  Rank-based contrastive loss incorporates the continuity of survival time through sample comparisons. Angle-based optimization preserves the hierarchical structure of hyperbolic space by optimizing the geographical constraints between modalities. Together, this module enables effective multimodal feature alignment and maintains the continuity of survival time.
Finally, we introduce an \textit{Censor-Conditioned Uncertainty Constraint} (CUC) that leverages the properties of hyperbolic space: samples closer to the origin have higher uncertainty, while samples farther from the origin have lower uncertainty \cite{chen2023hyperbolic}. This helps constrain censored and uncensored samples in survival prediction, addressing the challenge of uncertain samples and improving prediction performance. In conclusion, our framework captures complex survival data relationships and enhances survival predictions by effectively fusing multimodal features and preserving biological pathway insights through relative comparisons.

In summary, our contributions are summarized as follows: 
\begin{enumerate}
\raggedright
    \item  We present a novel HySurvPred, which represents the first work leveraging hyperbolic space to explore hierarchical intra-modal structures and inter-modal relationships in histopathology and genomics data.
    \item To improve multimodal feature representation in hyperbolic space, we propose the ARCL module, which employs ranking-based contrastive learning to capture the ordinal nature of survival time, and the CUC module, which effectively utilizes censored data based on uncertainty constraint.
    \item Extensive experiments on five benchmark datasets demonstrate significant improvements over state-of-the-art methods in multimodal survival prediction.
\end{enumerate}
\section{Related Works}
\noindent
For survival prediction tasks, there are two popular approaches. First, instance-based methods for histopathology focus on selecting representative instances from Whole-Slide Images (WSIs) and aggregating them for prediction. Second, multimodal fusion methods combine different data types to capture cross-modal interactions for improved prediction accuracy.
\subsection{Multiple Instance Learning for Histopathology}
Handling WSIs with their large dimensions poses significant challenges, making Multiple Instance Learning (MIL) a key approach in histopathology. MIL treats WSIs as collections of smaller patches, either selecting representative instances or converting them into low-dimensional representations, which are then aggregated to form a bag-level prediction.

Clustering-based methods \cite{sharma2021cluster,wang2019weakly} group instance embeddings into centroids, which are then used for the final prediction. Among these, attention-based MIL (AB-MIL) \cite{ilse2018attention,li2021dual} has gained prominence by assigning attention weights to instances. Different AB-MIL approaches vary in how these weights are calculated. For instance, an early AB-MIL model \cite{ilse2018attention} employed a side-branch network for weight generation, while DS-MIL \cite{li2021dual} utilized cosine similarity to assign weights.
Recently, transformer-based methods, such as TransMIL \cite{shao2021transmil}, have incorporated self-attention mechanisms to capture long-range interactions in WSIs. Moreover, DTFD-MIL \cite{zhang2022dtfd} introduced a double-tier MIL framework to refine instance weighting. Despite extensive exploration in previous work, many studies focus on unimodal data, neglecting to integrate different modalities. This limits the ability to fully explore the biological information relationships across modalities, which is crucial for more accurate survival analysis.

\subsection{Multimodal Fusion for Survival Prediction}
In clinical settings, patients are often assessed using diverse multimodal data, such as genomics \cite{klambauer2017self}, histopathology \cite{liu2024advmil,chen2022scaling}, and radiology \cite{yao2021deepprognosis}. These datasets play a critical role in diagnosis and prognosis, driving the need to explore effective methods for learning interactions across modalities \cite{zhang2023tformer}. Existing multimodal fusion approaches generally fall into two main categories: tensor-based and attention-based techniques \cite{zhang2020multimodal}.

Tensor-based methods, like simple concatenation or bilinear pooling, are efficient but limited in capturing complex interactions across modalities. These methods, such as Kronecker product \cite{wang2021gpdbn} or factorized bilinear pooling \cite{li2022hfbsurv}, are typically applied at early or late fusion stages, potentially missing inter-modal relationships \cite{chen2022pan}.
Attention-based methods, such as co-attention \cite{chen2021multimodal} or optimal transport \cite{xu2023multimodal}, have gained popularity for capturing cross-modal correlations. Despite their strength in reducing redundancy, these methods often struggle to preserve modality-specific information effectively. Despite these advancements, many studies still rely on generic survival functions and feature learning for survival prediction, failing to fully exploit the hierarchical information between histopathology and genomic data. Additionally, they overlook the continuity of survival time and ordinal risk and do not effectively optimize censored data, limiting their potential for accurate and nuanced survival analysis.
\section{Experiment and Results}
\noindent
In this section, we first outline the datasets and experimental settings, adhering to established protocols \cite{9710773} to ensure fair comparisons. Subsequently, we evaluate the performance of the proposed method against state-of-the-art (SOTA) techniques, encompassing both unimodal and multimodal approaches. We then analyze the contribution of each component in our method. Lastly, we provide Kaplan-Meier survival curve to evaluate survival analysis outcomes from a statistical perspective and t-SNE visualization of feature distribution.

\subsection{Datasets and Settings}
We utilized five publicly available cancer datasets from The Cancer Genome Atlas (TCGA) and conducted extensive experiments on these datasets. Each dataset includes paired diagnostic WSIs and genomic data with ground-truth survival outcomes. The number of cases for each cancer type is as follows: Bladder Urothelial Carcinoma (BLCA) with 373 cases, Breast Invasive Carcinoma (BRCA) with 956 cases, Uterine Corpus Endometrial Carcinoma (UCEC) with 480 cases, Glioblastoma and Lower Grade Glioma (GBMLGG) with 569 cases, and Lung Adenocarcinoma (LUAD) with 453 cases. It is worth noting that the number of cases used in all comparative methods is no fewer than the cases used in the proposed method. For genomic data, the number of unique functional categories ($M_g$) is set to 6, following \cite{LIBERZON2015417,9710773}, which include: (1) Tumor Suppression, (2) Oncogenesis, (3) Protein Kinases, (4) Cellular Differentiation, (5) Transcription, and (6) Cytokines and Growth.

\subsection{Comparisons With State-of-the-Arts}
To demonstrate the robustness of our framework, we evaluate our method against unimodal baselines and state-of-the-art (SOTA) multimodal approaches as shown in Table \ref{table1}. 
Compared to multimodal models, including MCAT \cite{chen2021multimodal}, Porpoise \cite{chen2022pan}, MOTCat \cite{xu2023multimodal}, CMTA \cite{zhou2023cross}, CGM \cite{zhou2024cohort}, SurvPath \cite{jaume2024modeling}, our model achieves mean values of 0.711, 0.757, 0.726, 0.709 and 0.859 C-index scores on BLCA, BRCA, UCEC, LUAD and GBMLGG, which represents state-of-the-art performance on most datasets. It is also the second-best on the GBMLGG dataset, with a small margin behind the best 0.861 reported by CGM \cite{zhou2024cohort}. Notably, while CGM performs well on the GBMLGG dataset, it still falls short compared to our method on the other datasets.
For unimodal models, the BLCA dataset performs better than the other datasets when only the histopathology modality is used in DTFDMIL \cite{zhang2022dtfd}, while the GBMLGG dataset performs better than the others when only the genomic modality is used in SNNTrans \cite{klambauer2017self}. However, the multimodal approach, overall, outperforms the unimodal approach, demonstrating that there is complementary information between the two modalities. Compared to unimodal approaches based on genomics or histopathology, our framework has achieved significant improvements across all datasets. This suggests that our framework can effectively integrate complementary information from genomic datasets and histopathology images. Besides, we conducted a t-test and found that the p-value was less than 0.05 across all datasets, indicating that our method significantly outperforms MOTCat, as shown in Fig. \ref{fig:KA}.

\subsection{Ablation Study}
\begin{figure*}[h!]
\centering
\includegraphics[width=1.0\linewidth]{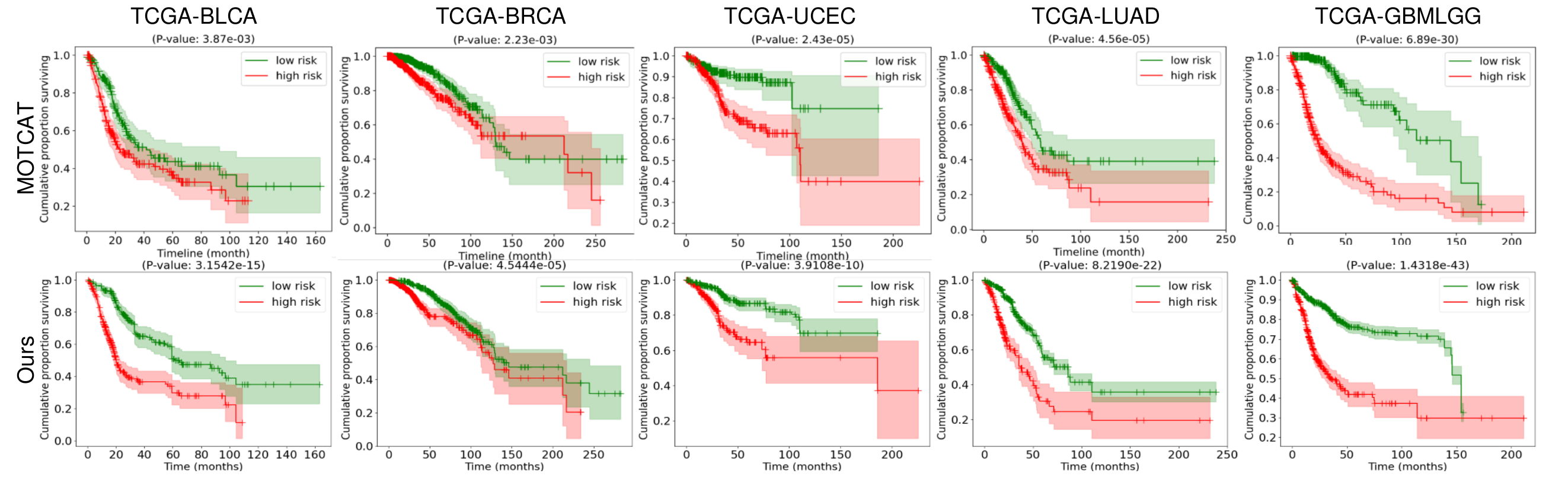}
\caption{ Kaplan-Meier Analysis on five cancer datasets, where patient stratifications of low risk (green) and high risk (red) are presented. Shaded areas refer to the confidence intervals. P-value $<$ 0.05 means a significant statistical difference between two groups, and a lower P-value is better.}
\label{fig:KA}
\vspace{-0.1cm}
\end{figure*}
To demonstrate the effectiveness of our method, we incrementally attach our modules to our baseline MOTCat, including MHM, ARCL and CUC. 
We first reimplement the MOTCat \cite{xu2023multimodal} work and the results are shown in the $\text{Baseline}^\ast$ row of Table \ref{tab:component}, and then we only added any module of MHM or ARCL for survival prediction. CUC can only appear as a regular term, it cannot be used as a separate contribution, so we compared the effects of adding CUC to the other two modules and without it.
In the following part, we will describe the ablation results of each module separately, statistical analysis for survival dataset and the T-SNE visualization of feature changes.

\textbf{C-index Comparison.} The specific results are shown in Table \ref{tab:component}, where we observe that the MHM module provides a modest improvement in the final prediction, while the ARCL module shows more significant effectiveness. Additionally, ablation experiments on the CUC demonstrate that adding regular terms improves survival prediction performance, as evidenced by better C-index results.

\textbf{Kaplan-Meier Analysis.} To demonstrate the statistical differences in patient stratification performance, we present Kaplan-Meier survival curves for various methods. To evaluate the statistical significance between these two groups, the Log-rank test is employed, where a smaller P-value indicates superior stratification performance. As shown in Fig. \ref{fig:KA}, our method consistently provides a clearer separation between low-risk and high-risk patients across all datasets. Furthermore, the Log-rank test results reveal that our approach achieves significantly lower P-values compared to our baseline MOTCat, particularly on BLCA, BRCA, and UCEC datasets, showing substantial improvements.

\textbf{T-SNE Visualization.}
We visualized the original multimodal features and the features after updates, as shown in Fig. 4. Initially, in hyperbolic space, the features of histopathology slides and genomic data are mixed together. However, after dynamic updates, the histopathology and genomic features become more separated, preserving the hierarchical characteristics of each modality and ensuring clearer distinctions between them.

\begin{figure}[ht]
    \centering
    \begin{minipage}{0.2\textwidth}
        \centering
        \includegraphics[width=\textwidth, height=0.53in]{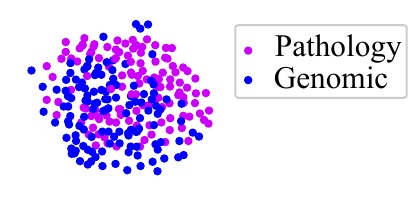}
        \caption*{(a)}
        \label{fig:image_a}
    \end{minipage}
    \hfill
    \begin{minipage}{0.2\textwidth}
        \centering
        \includegraphics[width=\textwidth, height=0.53in]{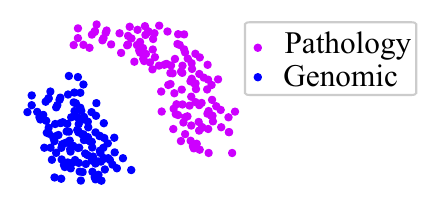}
        \caption*{(b)}
        \label{fig:image_b}
    \end{minipage}
    \caption{(a) Original hyperbolic features. (b) After hyperbolic feature updates.}
    \label{fig:t-sne1}
\end{figure}
\begin{figure}[ht]
    \centering
    \begin{minipage}{0.2\textwidth}
        \centering
        \includegraphics[width=\textwidth, height=0.74in]{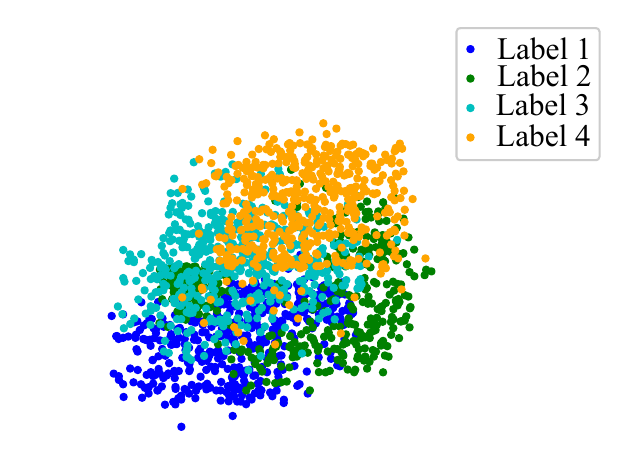}
        \caption*{(a)}
        \label{fig:image_a}
    \end{minipage}
    \hfill
    \begin{minipage}{0.2\textwidth}
        \centering
        \includegraphics[width=\textwidth, height=0.74in]{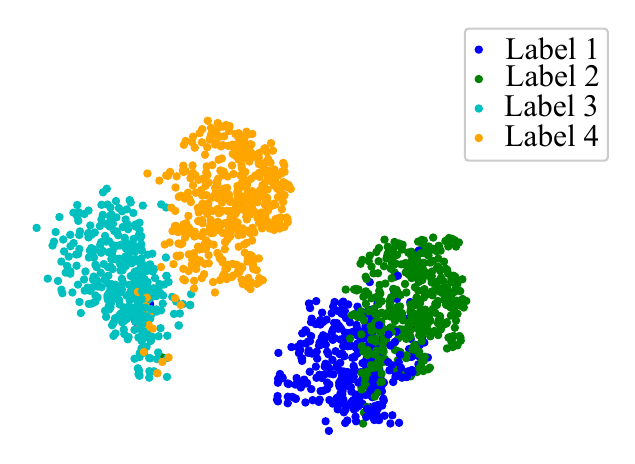}
        \caption*{(b)}
        \label{fig:image_b}
    \end{minipage}
    \caption{(a) Original risk level distribution. (b) Risk level distribution after updates.}
    \label{fig:t-sne1}
\end{figure}
Besides, we visualized the fused multimodal features according to risk levels. Fig. 5 shows features from different risk levels have achieved disentanglement during the update process, enabling clearer distinctions between risk categories.

\section{Conclusion}
\noindent
In this work, we present a novel framework, HySurvPred, to explore the inherent hierarchical structures within and across the histopathology and genomics modalities. 
To preserve the ordinal nature of survival time, our ARCL module introduces ranking-based contrastive learning for multimodal feature alignment. Motivated by the inherent property of hyperbolic space, our CUC module effectively utilizes censored data based on uncertainty constraint. This framework enhances the hierarchical representation of multimodal features, and highlights the continuity of survival time and the impact of censored data. The effectiveness of the HySurvPred framework and our proposed modules are validated on five benchmark datasets.

\bibliographystyle{named}
\bibliography{ijcai24}

\begin{thebibliography}{}

\bibitem[\protect\citeauthoryear{Chen \bgroup \em et al.\egroup }{2021a}]{chen2021multimodal}
Richard~J Chen, Ming~Y Lu, Wei-Hung Weng, Tiffany~Y Chen, Drew~FK Williamson, Trevor Manz, Maha Shady, and Faisal Mahmood.
\newblock Multimodal co-attention transformer for survival prediction in gigapixel whole slide images.
\newblock In {\em Proceedings of the IEEE Int. Conf. Comput. Vis.}, pages 4015--4025, 2021.

\bibitem[\protect\citeauthoryear{Chen \bgroup \em et al.\egroup }{2021b}]{9710773}
Richard~J. Chen, Ming~Y. Lu, Wei-Hung Weng, Tiffany~Y. Chen, Drew~FK. Williamson, Trevor Manz, Maha Shady, and Faisal Mahmood.
\newblock Multimodal co-attention transformer for survival prediction in gigapixel whole slide images.
\newblock In {\em Proc. IEEE Int. Conf. Comput. Vis.}, pages 3995--4005, 2021.

\bibitem[\protect\citeauthoryear{Chen \bgroup \em et al.\egroup }{2022a}]{chen2022scaling}
Richard~J Chen, Chengkuan Chen, Yicong Li, Tiffany~Y Chen, Andrew~D Trister, Rahul~G Krishnan, and Faisal Mahmood.
\newblock Scaling vision transformers to gigapixel images via hierarchical self-supervised learning.
\newblock In {\em Proceedings of the IEEE Conf. Comput. Vis. Pattern Recognit.}, pages 16144--16155, 2022.

\bibitem[\protect\citeauthoryear{Chen \bgroup \em et al.\egroup }{2022b}]{chen2022pan}
Richard~J Chen, Ming~Y Lu, Drew~FK Williamson, Tiffany~Y Chen, Jana Lipkova, Zahra Noor, Muhammad Shaban, Maha Shady, Mane Williams, Bumjin Joo, et~al.
\newblock Pan-cancer integrative histology-genomic analysis via multimodal deep learning.
\newblock {\em Cancer Cell}, 40(8):865--878, 2022.

\bibitem[\protect\citeauthoryear{Chen \bgroup \em et al.\egroup }{2023}]{chen2023hyperbolic}
Bike Chen, Wei Peng, Xiaofeng Cao, and Juha R{\"o}ning.
\newblock Hyperbolic uncertainty aware semantic segmentation.
\newblock {\em IEEE Transactions on Intelligent Transportation Systems}, 2023.

\bibitem[\protect\citeauthoryear{Ilse \bgroup \em et al.\egroup }{2018}]{ilse2018attention}
Maximilian Ilse, Jakub Tomczak, and Max Welling.
\newblock Attention-based deep multiple instance learning.
\newblock In {\em Proc. Int. Conf. Mach. Learn.}, pages 2127--2136. PMLR, 2018.

\bibitem[\protect\citeauthoryear{Jaume \bgroup \em et al.\egroup }{2024}]{jaume2024modeling}
Guillaume Jaume, Anurag Vaidya, Richard~J Chen, Drew~FK Williamson, Paul~Pu Liang, and Faisal Mahmood.
\newblock Modeling dense multimodal interactions between biological pathways and histology for survival prediction.
\newblock In {\em Proceedings of the IEEE Conf. Comput. Vis. Pattern Recognit.}, pages 11579--11590, 2024.

\bibitem[\protect\citeauthoryear{Klambauer \bgroup \em et al.\egroup }{2017}]{klambauer2017self}
G{\"u}nter Klambauer, Thomas Unterthiner, Andreas Mayr, and Sepp Hochreiter.
\newblock Self-normalizing neural networks.
\newblock {\em Proc. Adv. Neural Inf. Process. Syst.}, 30, 2017.

\bibitem[\protect\citeauthoryear{Kuroda \bgroup \em et al.\egroup }{2021}]{kuroda2021tumor}
Hajime Kuroda, Tsengelmaa Jamiyan, Rin Yamaguchi, Akinari Kakumoto, Akihito Abe, Oi~Harada, and Atsuko Masunaga.
\newblock Tumor-infiltrating b cells and t cells correlate with postoperative prognosis in triple-negative carcinoma of the breast.
\newblock {\em BMC cancer}, 21:1--10, 2021.

\bibitem[\protect\citeauthoryear{Li \bgroup \em et al.\egroup }{2021}]{li2021dual}
Bin Li, Yin Li, and Kevin~W Eliceiri.
\newblock Dual-stream multiple instance learning network for whole slide image classification with self-supervised contrastive learning.
\newblock In {\em Proceedings of the IEEE Conf. Comput. Vis. Pattern Recognit.}, pages 14318--14328, 2021.

\bibitem[\protect\citeauthoryear{Li \bgroup \em et al.\egroup }{2022}]{li2022hfbsurv}
Ruiqing Li, Xingqi Wu, Ao~Li, and Minghui Wang.
\newblock Hfbsurv: hierarchical multimodal fusion with factorized bilinear models for cancer survival prediction.
\newblock {\em Bioinformatics}, 38(9):2587--2594, 2022.

\bibitem[\protect\citeauthoryear{Liberzon \bgroup \em et al.\egroup }{2015}]{LIBERZON2015417}
Arthur Liberzon, Chet Birger, Helga Thorvaldsdóttir, Mahmoud Ghandi, Jill P. Mesirov, and Pablo Tamayo.
\newblock The molecular signatures database hallmark gene set collection.
\newblock {\em CELL SYST}, 1(6):417--425, 2015.

\bibitem[\protect\citeauthoryear{Liu \bgroup \em et al.\egroup }{2024}]{liu2024advmil}
Pei Liu, Luping Ji, Feng Ye, and Bo~Fu.
\newblock Advmil: Adversarial multiple instance learning for the survival analysis on whole-slide images.
\newblock {\em MED IMAGE ANAL}, 91:103020, 2024.

\bibitem[\protect\citeauthoryear{Oya \bgroup \em et al.\egroup }{2020}]{oya2020tumor}
Yukiko Oya, Yoku Hayakawa, and Kazuhiko Koike.
\newblock Tumor microenvironment in gastric cancers.
\newblock {\em Cancer science}, 111(8):2696--2707, 2020.

\bibitem[\protect\citeauthoryear{Qiu \bgroup \em et al.\egroup }{2023}]{qiu2023deep}
Lin Qiu, Aminollah Khormali, and Kai Liu.
\newblock Deep biological pathway informed pathology-genomic multimodal survival prediction.
\newblock {\em arXiv preprint arXiv:2301.02383}, 2023.

\bibitem[\protect\citeauthoryear{Shao \bgroup \em et al.\egroup }{2021}]{shao2021transmil}
Zhuchen Shao, Hao Bian, Yang Chen, Yifeng Wang, Jian Zhang, Xiangyang Ji, et~al.
\newblock Transmil: Transformer based correlated multiple instance learning for whole slide image classification.
\newblock {\em Proc. Adv. Neural Inf. Process. Syst.}, 34:2136--2147, 2021.

\bibitem[\protect\citeauthoryear{Sharma \bgroup \em et al.\egroup }{2021}]{sharma2021cluster}
Yash Sharma, Aman Shrivastava, Lubaina Ehsan, Christopher~A Moskaluk, Sana Syed, and Donald Brown.
\newblock Cluster-to-conquer: A framework for end-to-end multi-instance learning for whole slide image classification.
\newblock In {\em Proc. Int. Conf. Medical Imaging Deep Learn.}, pages 682--698. PMLR, 2021.

\bibitem[\protect\citeauthoryear{Song \bgroup \em et al.\egroup }{2024a}]{song2024morphological}
Andrew~H Song, Richard~J Chen, Tong Ding, Drew~FK Williamson, Guillaume Jaume, and Faisal Mahmood.
\newblock Morphological prototyping for unsupervised slide representation learning in computational pathology.
\newblock In {\em Proceedings of the IEEE Conf. Comput. Vis. Pattern Recognit.}, pages 11566--11578, 2024.

\bibitem[\protect\citeauthoryear{Song \bgroup \em et al.\egroup }{2024b}]{song2024analysis}
Andrew~H Song, Mane Williams, Drew~FK Williamson, Sarah~SL Chow, Guillaume Jaume, Gan Gao, Andrew Zhang, Bowen Chen, Alexander~S Baras, Robert Serafin, et~al.
\newblock Analysis of 3d pathology samples using weakly supervised ai.
\newblock {\em Cell}, 187(10):2502--2520, 2024.

\bibitem[\protect\citeauthoryear{Wang \bgroup \em et al.\egroup }{2019}]{wang2019weakly}
Xi~Wang, Hao Chen, Caixia Gan, Huangjing Lin, Qi~Dou, Efstratios Tsougenis, Qitao Huang, Muyan Cai, and Pheng-Ann Heng.
\newblock Weakly supervised deep learning for whole slide lung cancer image analysis.
\newblock {\em IEEE T CYBERNETICS}, 50(9):3950--3962, 2019.

\bibitem[\protect\citeauthoryear{Wang \bgroup \em et al.\egroup }{2021}]{wang2021gpdbn}
Zhiqin Wang, Ruiqing Li, Minghui Wang, and Ao~Li.
\newblock Gpdbn: deep bilinear network integrating both genomic data and pathological images for breast cancer prognosis prediction.
\newblock {\em Bioinformatics}, 37(18):2963--2970, 2021.

\bibitem[\protect\citeauthoryear{Xu and Chen}{2023}]{xu2023multimodal}
Yingxue Xu and Hao Chen.
\newblock Multimodal optimal transport-based co-attention transformer with global structure consistency for survival prediction.
\newblock In {\em Proceedings of the IEEE Int. Conf. Comput. Vis.}, pages 21241--21251, 2023.

\bibitem[\protect\citeauthoryear{Yao \bgroup \em et al.\egroup }{2021}]{yao2021deepprognosis}
Jiawen Yao, Yu~Shi, Kai Cao, Le~Lu, Jianping Lu, Qike Song, Gang Jin, Jing Xiao, Yang Hou, and Ling Zhang.
\newblock Deepprognosis: Preoperative prediction of pancreatic cancer survival and surgical margin via comprehensive understanding of dynamic contrast-enhanced ct imaging and tumor-vascular contact parsing.
\newblock {\em MED IMAGE ANAL}, 73:102150, 2021.

\bibitem[\protect\citeauthoryear{Zadeh and Schmid}{2020}]{zadeh2020bias}
Shekoufeh~Gorgi Zadeh and Matthias Schmid.
\newblock Bias in cross-entropy-based training of deep survival networks.
\newblock {\em IEEE T PATTERN ANAL}, 43(9):3126--3137, 2020.

\bibitem[\protect\citeauthoryear{Zhang \bgroup \em et al.\egroup }{2020a}]{zhang2020multimodal}
Chao Zhang, Zichao Yang, Xiaodong He, and Li~Deng.
\newblock Multimodal intelligence: Representation learning, information fusion, and applications.
\newblock {\em IEEE J-STSP}, 14(3):478--493, 2020.

\bibitem[\protect\citeauthoryear{Zhang \bgroup \em et al.\egroup }{2020b}]{zhang2020evaluation}
Renyu Zhang, Aly~A Khan, and Robert~L Grossman.
\newblock Evaluation of hyperbolic attention in histopathology images.
\newblock In {\em 2020 IEEE 20th International Conference on Bioinformatics and Bioengineering (BIBE)}, pages 773--776. IEEE, 2020.

\bibitem[\protect\citeauthoryear{Zhang \bgroup \em et al.\egroup }{2022}]{zhang2022dtfd}
Hongrun Zhang, Yanda Meng, Yitian Zhao, Yihong Qiao, Xiaoyun Yang, Sarah~E Coupland, and Yalin Zheng.
\newblock Dtfd-mil: Double-tier feature distillation multiple instance learning for histopathology whole slide image classification.
\newblock In {\em Proceedings of the IEEE Conf. Comput. Vis. Pattern Recognit.}, pages 18802--18812, 2022.

\bibitem[\protect\citeauthoryear{Zhang \bgroup \em et al.\egroup }{2023}]{zhang2023tformer}
Yilan Zhang, Fengying Xie, and Jianqi Chen.
\newblock Tformer: A throughout fusion transformer for multi-modal skin lesion diagnosis.
\newblock {\em Comput Biol Med}, 157:106712, 2023.

\bibitem[\protect\citeauthoryear{Zhou and Chen}{2023}]{zhou2023cross}
Fengtao Zhou and Hao Chen.
\newblock Cross-modal translation and alignment for survival analysis.
\newblock In {\em Proceedings of the IEEE Int. Conf. Comput. Vis.}, pages 21485--21494, 2023.

\bibitem[\protect\citeauthoryear{Zhou and Sharpee}{2021}]{zhou2021hyperbolic}
Yuansheng Zhou and Tatyana~O Sharpee.
\newblock Hyperbolic geometry of gene expression.
\newblock {\em Iscience}, 24(3), 2021.

\bibitem[\protect\citeauthoryear{Zhou \bgroup \em et al.\egroup }{2024}]{zhou2024cohort}
Huajun Zhou, Fengtao Zhou, and Hao Chen.
\newblock Cohort-individual cooperative learning for multimodal cancer survival analysis.
\newblock {\em arXiv preprint arXiv:2404.02394}, 2024.

\bibitem[\protect\citeauthoryear{Zhu \bgroup \em et al.\egroup }{2017}]{zhu2017wsisa}
Xinliang Zhu, Jiawen Yao, Feiyun Zhu, and Junzhou Huang.
\newblock Wsisa: Making survival prediction from whole slide histopathological images.
\newblock In {\em Proc. IEEE Conf. Comput. Vis. Pattern Recognit.}, pages 7234--7242, 2017.

\bibitem[\protect\citeauthoryear{Zuo \bgroup \em et al.\egroup }{2022}]{zuo2022identify}
Yingli Zuo, Yawen Wu, Zixiao Lu, Qi~Zhu, Kun Huang, Daoqiang Zhang, and Wei Shao.
\newblock Identify consistent imaging genomic biomarkers for characterizing the survival-associated interactions between tumor-infiltrating lymphocytes and tumors.
\newblock In {\em Proc. Int. Conf. Med. Image Comput. Comput.-Assisted Intervention}, pages 222--231. Springer, 2022.

\end{thebibliography}

\end{document}